\documentclass[lettersize,journal]{IEEEtran}  

\usepackage{epsfig} 
\usepackage{mathptmx} 
\usepackage{times} 
\usepackage{amsmath} 
\usepackage{amssymb}  
\usepackage[caption=false,font=normalsize,labelfont=sf,textfont=sf]{subfig}
\usepackage{booktabs}
\usepackage{multirow}
\usepackage{graphicx}
\usepackage{float} 
\usepackage{pdflscape} 
\usepackage{tabularx} 
\usepackage{booktabs}
\usepackage{multirow}
\usepackage{colortbl}
\usepackage{bbding}
\usepackage{algorithm}
\usepackage{pifont}
\usepackage{wasysym}
\usepackage{utfsym}

\usepackage{hyperref}
\begin{document}
\title{DSM: Constructing a Diverse Semantic Map for 3D Visual Grounding
}

\author{Qinghongbing Xie$^{1}$$^{\dag}$, Zijian Liang$^{2}$$^{\dag}$, Fuhao Li and Long Zeng$^{1}$$^{*}$
\\
Project Page: \href{https://binicey.github.io/DSM/}{https://binicey.github.io/DSM}
\thanks{$^{\dag}$Equal contribution.}
\thanks{$^{*}$Corresponding author. (E-mail: zenglong@sz.tsinghua.edu.cn)}%
\thanks{$^{1}$Qinghongbing Xie, Zijian Liang, Fuhao Li, and Long Zeng are with Tsinghua Shenzhen International
 Graduate School, Tsinghua University, Shenzhen, China.}
}


\markboth{Journal of \LaTeX\ Class Files,~Vol.~x, No.~x, x~x}%
{Shell \MakeLowercase{\textit{et al.}}: A Sample Article Using IEEEtran.cls for IEEE Journals}


\maketitle

\begin{abstract}

Effective scene representation is critical for the visual grounding ability of representations, yet existing methods for 3D Visual Grounding are often constrained. They either only focus on geometric and visual cues, or, like traditional 3D scene graphs, lack the multi-dimensional attributes needed for complex reasoning. To bridge this gap, we introduce the Diverse Semantic Map (DSM) framework, a novel scene representation framework that enriches robust geometric models with a spectrum of VLM-derived semantics, including appearance, physical properties, and affordances. The DSM is first constructed online by fusing multi-view observations within a temporal sliding window, creating a persistent and comprehensive world model. Building on this foundation, we propose DSM-Grounding, a new paradigm that shifts grounding from free-form VLM queries to a structured reasoning process over the semantic-rich map, markedly improving accuracy and interpretability. Extensive evaluations validate our approach's superiority. On the ScanRefer benchmark, DSM-Grounding achieves a state-of-the-art 59.06\% overall accuracy of IoU@0.5, surpassing others by 10\%. In semantic segmentation, our DSM attains a 67.93\% F-mIoU, outperforming all baselines, including privileged ones. Furthermore, successful deployment on physical robots for complex navigation and grasping tasks confirms the framework's practical utility in real-world scenarios.


\end{abstract}

\section{Introduction}
Effective scene representation is a cornerstone for robotic agents to operate robustly in real-world environments.\cite{mascaro2024scene} It provides the essential foundation for environmental perception and interpretation, which is particularly critical for complex tasks like 3D Visual Grounding. For example, a service robot instructed to retrieve a specific fruit must construct a mental model that encodes not only the fruit's identity but also its intricate spatial and semantic relationships with surrounding entities, such as the refrigerator it is in.

However, prior research in 3D Visual Grounding has predominantly only focused on geometric and visual cues, like view selection optimization.\cite{li2024seeground,zhu2024scanreason} While these advancements are valuable, they often neglect the rich intrinsic attributes of objects and their contextual interdependencies. This oversight limits the robot's ability to perform advanced reasoning, as it fails to leverage the implicit semantic and physical logic inherent in the scene. Real-world scenes contain diverse contextual information, with attributes like an apple's color, freshness, weight, and position being critical for robotic tasks. An effective scene representation must be both expressive, encoding rich information, and compact, ensuring adaptability across various robotic platforms. Existing 3D scene graphs, however, only focus on simple semantics when capturing the attributes within a scene \cite{rosinol20203d, wu2021scenegraphfusion}, making it difficult to support the reasoning of large models in complex environments.

\begin{figure}[H]
    \centering
    \includegraphics[trim=260 200 240 180, clip, width=0.48\textwidth]{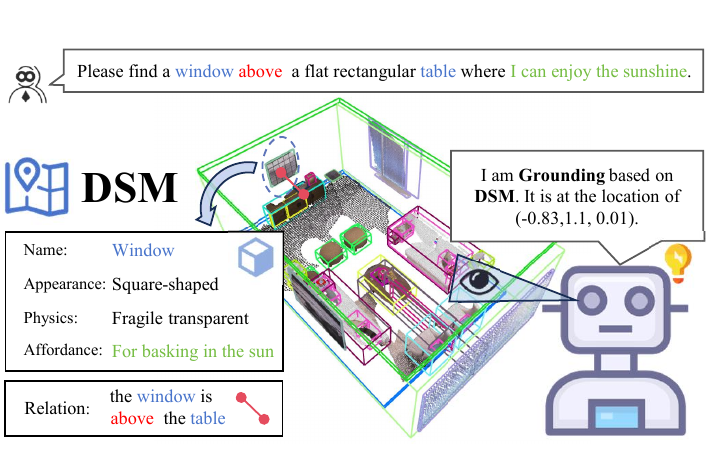}
    \caption{Our work introduces a novel scene representation, the Diverse Semantic Map (DSM) framework, designed to enhance deep reasoning in the 3D Visual Grounding task.}
    \label{fig:Motivation}
\end{figure}

To address this duality, we propose the Diverse Semantic Map (DSM) framework, a novel scene representation framework designed for 3D Visual Grounding that systematically incorporates multi-dimensional object attributes and their interrelationships. Our method leverages Vision-Language Models (VLMs) to construct this map, which populates the scene with not only geometric features but also a spectrum of rich semantic attributes, including appearance, physical properties, and affordances. By providing a richer, more nuanced scene understanding, the DSM framework enhances the adaptability and effectiveness of robotic systems.

Robots perceive their environment through a continuous stream of first-person observations. This temporal data is crucial, as it offers multi-view perspectives that naturally resolve geometric ambiguities and uncover latent semantic features. Capitalizing on this principle, we introduce a novel time-window-based construction method. This method systematically extracts and fuses multi-view semantic information from the temporal data stream to build the DSM, populating it with robust geometric models and comprehensive attribute profiles for each object instance.

Existing VLMs often struggle with the complex spatial and relational reasoning required for precise grounding, and are limited by their input formats, only being able to interpret simple scene graphs.\cite{gu2024conceptgraphs,maggio2024clio} To address this, we propose DSM-Grounding, a novel grounding method specifically designed to leverage the structured, multi-faceted information within the DSM. This approach transforms the grounding task from a direct VLM query to a structured search and reasoning process over the semantic map, thereby enhancing both accuracy and interpretability.

Our empirical evaluations demonstrate that DSM-Grounding significantly outperforms state-of-the-art methods in 3D Visual Grounding Tasks. For instance, on the challenging ScanRefer benchmark, our method achieves an overall accuracy of 59.06 at an IoU threshold of 0.5, surpassing prior zero-shot approaches by a significant margin. Furthermore, we showcase the DSM framework's versatility and practical utility by successfully deploying it in downstream robotics tasks, including navigation and grasping. These experiments validate its robust performance in complex, interactive scenarios.

In summary, this paper introduces three core contributions to improve understanding of the 3D scene for robotic agents.

\textbf{Our main contributions are as follows:} 
\begin{enumerate}
    \item We propose the Diverse Semantic Map (DSM) framework, which is capable of supporting complex multi-dimensional scene representation and enabling grounding that integrates both semantic understanding and precise localization.
    \item We develop a time-window-based mapping method that integrates geometric and semantic perception, and construct a DSM component to represent the rich semantics within a scene.
    \item We present DSM-Grounding, a new 3D grounding method that leverages the DSM to enable deeper scene reasoning for robotic agents.
\end{enumerate}

\section{Related Work}
\textbf{3D Scene Representation}
Effective 3D scene representation is fundamental for robot autonomy, evolving from purely geometric maps to rich, semantic structures \cite{mascaro2024scene}. Early methods focused on metric-semantic mapping, augmenting geometric reconstructions with object-level labels. Systems like Kimera \cite{rosinol2020kimera} pioneered real-time, dense semantic mesh generation. The advent of foundation models has enabled open-vocabulary mapping, with methods like ConceptFusion \cite{conceptfusion} creating language-grounded 3D maps without predefined categories. To capture more complex environmental structure, research has advanced towards 3D Scene Graphs (3DSGs), which explicitly model objects and their interrelationships. Works like SceneGraphFusion \cite{wu2021scenegraphfusion} and Hydra \cite{hughes2022hydra} build hierarchical representations of scenes. More recently, ConceptGraphs \cite{gu2024conceptgraphs} and Clio \cite{maggio2024clio} have leveraged Large Language Models (LLMs) to construct open-vocabulary 3DSGs. However, these representations often focus on categorical labels and spatial relations, lacking the fine-grained, multi-dimensional attributes (e.g., physical properties, affordances) required for complex reasoning tasks. Our DSM addresses this gap by creating a more expressive and diverse semantic map.

\textbf{3D Scene Graph}
3D Scene Graphs (3DSGs) offer a structured and symbolic representation of environments, capturing objects as nodes and their relationships as edges, which is invaluable for high-level reasoning \cite{bae2022survey}. Early approaches were often supervised, relying on predefined object and relation categories \cite{rosinol20203d, wu2021scenegraphfusion}. The recent integration of LLMs has spurred the development of zero-shot, open-vocabulary 3DSG construction \cite{werby2024hierarchical, gu2024conceptgraphs, changcontext}, which leverages the models' general world knowledge. Despite these advances, current 3DSGs are often limited to describing explicit spatial or simple semantic connections (e.g., \textit{chair next to table}). They fall short in capturing the rich, implicit attributes—such as an object's material, weight, or intended use (affordance). Our work extends beyond this paradigm by explicitly modeling these diverse semantic dimensions.

\textbf{3D Visual Grounding}
3D Visual Grounding aims to localize objects in a 3D scene based on natural language descriptions. This task is central to human-robot communication. Open-vocabulary methods like OpenScene \cite{peng2023openscene}, NuGrounding\cite{li2025nugrounding} and Open3DIS \cite{nguyen2024open3dis} achieve grounding by aligning text and visual features in a shared embedding space. More recent approaches leverage the reasoning capabilities of VLMs. For instance, SeeGround \cite{li2024seeground} and ScanReason \cite{zhu2024scanreason} employ a render-and-prompt strategy, generating multiple views of candidate objects to query a VLM for the final decision. While effective, these methods treat the scene as a static entity to be queried. They often lack a persistent, structured world model, forcing them to reason from scratch with each new query and limiting their ability to leverage rich, pre-compiled semantic context. Our DSM-Grounding method overcomes this by transforming the task from a direct VLM query into a structured search and reasoning process over a persistent and semantically rich map, enhancing both efficiency and contextual understanding.

\section{Method}
\begin{figure*}[ht!]
    \centering
    \includegraphics[trim=25  90 25 70, clip, width=\textwidth]{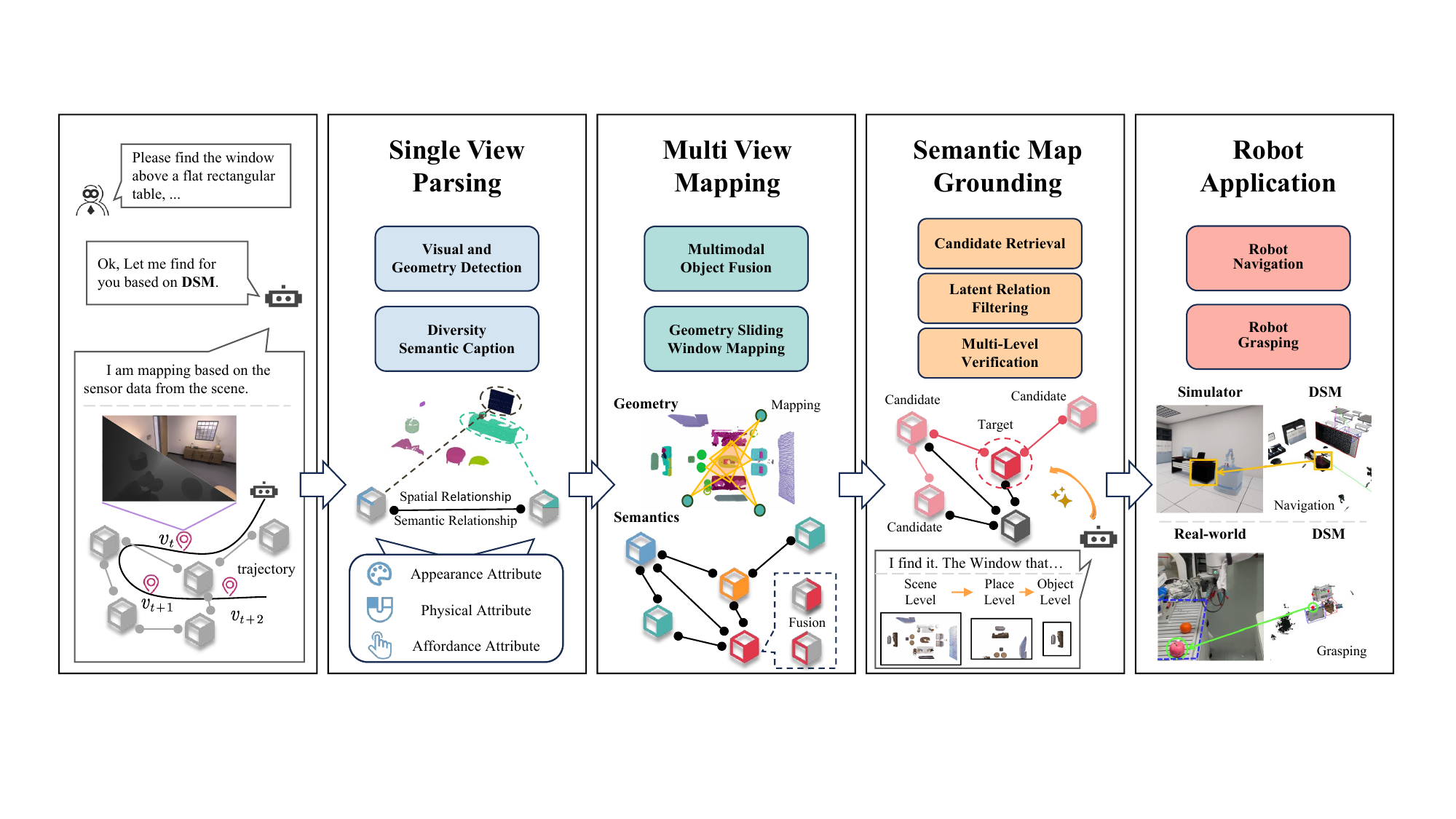}
    \caption{\textbf{Overview of the DSM framework}. After receiving the user's query, the robot first collects time-continuous poses, depth images, and color images of the scene to build a DSM. Next, we extract the visual and geometric information from each observation point. At the same time, we use VLM to analyze their relationships and semantic attributes, which are categorized into Appearance, Physical and Affordance Attributes. We fuse objects from multi views using a multimodal object fusion method in conjunction with the Geometry Sliding Window method for mapping. Finally, we identify candidates in the DSM based on the attributes and relationships of objects. We use the multi-level observations method to precisely locate the target object. Additionally, our method can be broadly applied to tasks such as robotic semantic navigation and semantic grasping.
}
    \label{fig:pipeline}
\end{figure*}

Our methodology is designed to equip robotic agents with a deep, contextual understanding of 3D environments for robust visual grounding. The core of our approach is the online construction of a Diverse Semantic Map (DSM), a novel scene representation that captures not only geometry but also a rich tapestry of multi-dimensional semantic attributes and inter-object relationships. The map serves as a structured representation for DSM-Grounding, transforming direct VLM queries into structured reasoning processes. As illustrated in Figure \ref{fig:pipeline}, our pipeline consists of two main stages: (1) construction of the DSM from a continuous stream of RGB-D observations including single-view parsing and Multi View Mapping, (2) leveraging the structured knowledge within the DSM to perform accurate and interpretable 3D visual grounding.

\subsection{Definition of Diverse Semantic Map (DSM)}

We define the Diverse Semantic Map (DSM) as a 3D Scene Graph $G = (\mathcal{O}, \mathcal{R})$, where $\mathcal{O}$ is a set of object nodes and $\mathcal{R}$ is a set of edges representing their relationships. Each object node $O_i \in \mathcal{O}$ encapsulates both geometric and semantic information.

\textbf{Geometric Representation ($O_i^g$)}: This component models the object's physical presence in the world, containing its 3D point cloud $P_i$, and an oriented bounding box $B_i$.

\textbf{Semantic Representation ($O_i^s$)}: This component stores a rich, multi-faceted description of the object, including:
Identity ($N_i^i$): A class label or name (e.g., \textit{chair}).
Attributes ($A_i$): A structured set of VLM-derived descriptions categorized into, Appearance ($a_a$), Physical ($a_p$), and Affordance ($a_o$).
Each edge $R_{ij} \in \mathcal{R}$ connects two objects ($O_i, O_j$) and is also categorized into: Spatial Relations ($R_{ij}^g$): Geometric relationships like next to, on top of. Semantic Relations ($R_{ij}^s$): Functional or compositional relationships like part of, used with.

\subsection{DSM Construction}
\textbf{Single View Parsing}. For each incoming RGB-D frame, we perform open-vocabulary object detection and segmentation to obtain precise 2D masks. These masks are then back-projected using depth and camera pose information to generate object-centric point clouds, which serve as the initial geometric evidence for each detected object.

We leverage a Vision-Language Model (VLM) to extract a rich semantic profile for each object. Using visual prompting and Chain-of-Thought (CoT) reasoning, the VLM generates structured textual descriptions across three key dimensions: 
\begin{itemize}

\item \textbf{Appearance attribute} \( a_a\): Describes the visual characteristics of objects, including color, patterns, and texture.
\item \textbf{Physical attribute} \( a_p\): Captures the physical properties of objects, such as weight, material composition, and surface smoothness.
\item  \textbf{Affordance attribute} \( a_o\): Defines the functional aspects, applications, and operational methods associated with objects.
\end{itemize}

Furthermore, the VLM describes the \textbf{Semantic Relationships} (e.g., functional, compositional) between co-visible objects and we extracts the Spatial Relationship from object's points cloud$P_i$. This process yields a comprehensive set of geometric and semantic data for each object from a single viewpoint, as exemplified in Table \ref{tab:example opf SA} and Table \ref{tab:expample of relation}.

\begin{table*}[h] 
    \centering
    \caption{Example of Semantic Attribute}
    \label{tab:example opf SA}
\begin{tabular}{@{}cccc@{}}
\toprule
\textbf{Name} & \textbf{Appearance Attribute}                                                                                                                                                                                                   & \textbf{Physical Attribute}                                                                                                                                                                                                                                                            & \textbf{Affordance Attribute}                                                                                                                                                                                                                                 \\ \midrule
\textbf{pillow}        & a soft, square pillow with a floral design                                                                                                                                                                                      & \begin{tabular}[c]{@{}c@{}}filled with a soft material, \\ providing compressibility and comfort\end{tabular}                                                                                                                                                                          & \begin{tabular}[c]{@{}c@{}}intended for support when sitting or lying down,\\  enhancing comfort in seating areas\end{tabular}                                                                                                                                \\ \midrule
\textbf{stool}         & \begin{tabular}[c]{@{}c@{}}A small, rounded seat with a padded top, \\ typically covered in a beige fabric.\\  The design is simple yet stylish, \\ featuring a soft cushion\\  that provides comfort for sitting.\end{tabular} & \begin{tabular}[c]{@{}c@{}}The stool is sturdy and stable,  designed to \\ support a person's weight effectively. \\ It is lightweight, allowing \\ for easy movement and positioning. \\ It can be used as a seating solution \\ or as a footrest due to its low profile.\end{tabular} & \begin{tabular}[c]{@{}c@{}}The stool serves primarily as a seating option \\ but can also be used as a footrest. \\ Additionally, its design allows it to \\ function as a small table when needed,\\  making it a versatile piece of furniture.\end{tabular} \\ \bottomrule
\end{tabular}
\end{table*}

\begin{table}[h] 
    \centering
        \caption{Example of Relation}
    \label{tab:expample of relation}
\begin{tabular}{@{}cccc@{}}
\toprule

\textbf{Object type}    & \textbf{Name}           & \textbf{Spatial Relation} & \textbf{Semantic Relation}                                                                                                                                         \\ \midrule
\multirow{2}{*}{Target} & \multirow{2}{*}{pillow} & \multirow{4}{*}{close by} & \multirow{4}{*}{\begin{tabular}[c]{@{}c@{}}The pillow is an accessory\\  placed on the sofa \\ for comfort and support \\ while sitting or lounging.\end{tabular}} \\
                        &                         &                           &                                                                                                                                                                    \\ \cmidrule(r){1-2}
\multirow{2}{*}{Anchor} & \multirow{2}{*}{sofa}   &                           &                                                                                                                                                                    \\
                        &                         &                           &                                                                                                                                                                    \\ \bottomrule
\end{tabular}

\end{table}
\textbf{Multi View Mapping}. To build a persistent and globally consistent map, observations from new frames must be associated with and fused into the existing DSM. This process involves two coupled steps: multimodal data association and map update.  For a newly observed object $O_{new}$ and an existing map object $O_i \in \mathcal{O}$, we compute a weighted multimodal similarity score to determine if they represent the same entity:

\begin{equation}
S = s_v + s_g + s_c
\end{equation}
\begin{equation}
s_v = \text{CosSimilarity}(f_{v\hat{p}}, f_{v\hat{q}}) 
\end{equation}
\begin{equation}
s_g = 
\begin{cases} 
s_{g0} & \text{if } bbox_p \text{ inside }  bbox_q\\
\text{IoU}(bbox_p, bbox_q) & \text{otherwise}
\end{cases}
\end{equation}
\begin{equation}
s_c = \text{CosSimilarity}(f_{s\hat{p}}, f_{s\hat{q}})  
\end{equation}

where \(s_v, s_g, s_s\) are the visual (embedding cosine similarity), geometric (3D IoU), and semantic (text embedding cosine similarity) scores, respectively. When two objects are mutually contained, the geometric score is set to a fixed value  \(s_{g0}\). If the total score exceeds a threshold, the objects are associated. \(f_{v\hat{p}}, f_{v\hat{q}}, f_{s\hat{p}}\), and \(f_{s\hat{q}}\) are the encoder features extracted from the objects' images and linguistic descriptions, respectively. 

\begin{figure}[H]
    \centering
    \includegraphics[trim=270  220 270 185, clip, width=0.5\textwidth]{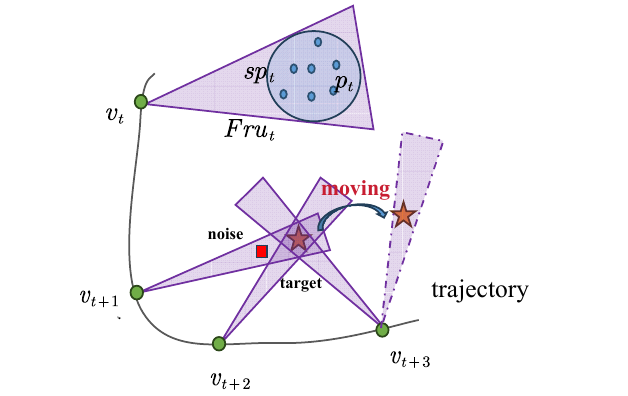}
    \caption{\textbf{Geometry Sliding Window Method.}We employ the Monte Carlo sampling method to estimate the observation frustum and subsequently optimize the object point cloud using multiple temporally continuous observation perspectives.}
    \label{fig:geometrySW}
\end{figure}

Upon successful association, we update the map object's geometric and semantic profiles. The geometry is refined using our Geometry Sliding Window Method, illustrated in Figure \ref{fig:geometrySW}. This method aggregates recent point cloud observations by constructing a viewing frustum for each frame, enabling robust noise filtering and shape completion. A spatial voting scheme is then applied to the aggregated point cloud, retaining points consistent with multiple views to filter noise and complete the shape. Concurrently, the object's semantic attributes and relations are updated via an aggregation and voting mechanism, reinforcing consistent information and ensuring the DSM becomes more accurate and robust over time.

\subsection{DSM-Grounding}

Our DSM-Grounding algorithm transforms the 3D visual grounding task from a direct VLM query on raw sensor data into a structured search and reasoning process over the pre-built DSM. This paradigm shift allows for deeper reasoning by leveraging the rich, multi-faceted information stored within the map. Given a natural language query $Q$, the process unfolds in three main stages: (1) Candidate Retrieval, (2) Latent Relation Filtering, and (3) Multi-Level Verification.

\textbf{Candidate Retrieval}. We use an LLM to parse the natural language query $Q$ into a structured format, identifying the primary target entity, any mentioned anchor entities, and their associated descriptive attributes. We then retrieve an initial set of candidate objects $\mathcal{O}_{cand} \subseteq \mathcal{O}$ from the DSM by matching these parsed entities against the object identities ($N_i$) and multi-dimensional attributes ($A_i$) stored in the map, using a combination of text matching techniques.

\textbf{Latent Relation Filtering(LRF)}. This module prunes the candidate set by verifying the relational constraints described in $Q$. For each candidate target $O_c \in \mathcal{O}{cand}$ and its potential anchor objects, we query the DSM for their stored relationship $\mathcal{R}$. We then use an LLM to score the consistency between the stored relationships and the relationship described in the query. LLM selects the Top-k candidates, resulting in a refined set $\mathcal{O}_{filtered}$. This step effectively leverages the pre-compiled relational knowledge in the DSM to resolve ambiguities.

\begin{figure}[h]
    \centering
    \includegraphics[trim=10  40 10 40,width=0.45\textwidth]{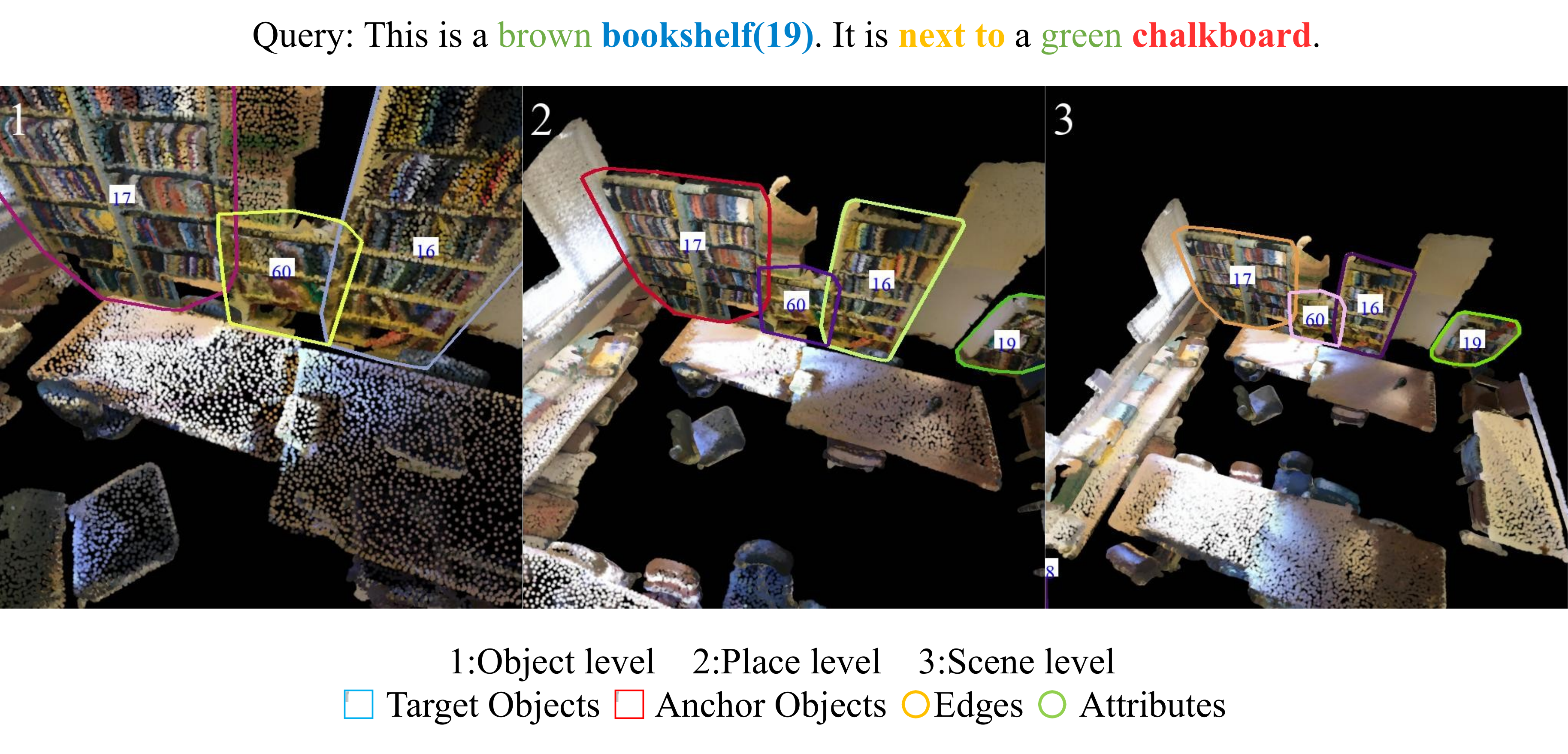}
    \caption{Multi-Level Observation.}
    \label{fig:render}
\end{figure}

\textbf{Multi-Level Verification}. For the final, small set of high-potential candidates in $\mathcal{O}_{filtered}$, we render images of each candidate from three perspectives: Object Level, Place Level, and Scene Level, leveraging DSM's geometric data for accurate visualization, as Fig \ref{fig:render}. 
\begin{itemize}
\item \textbf{Object Level}: the object fills the frame, providing detailed insight into its categories and attributes. 
\item \textbf{Place Level}: a broader view showing the relationship of objects with adjacent regions. 
\item \textbf{Scene Level}: the view is expanded to include almost the entire scene for contextual global information. 
\end{itemize}

These rendered views are presented to a VLM along with the object's rich semantic profile ($A_i$) retrieved from the DSM and the original query $Q$. The VLM then makes the final decision by reasoning over this high-quality, curated data, as in Eq.\ref{eq:grounding}. This final step uses the VLM's powerful reasoning ability not on the noisy, raw scene, but on focused, context-rich information pre-processed and structured by our DSM.

\begin{equation}
\label{eq:grounding}
\mathrm{pred} = \mathrm{VLM}\left(I,O_{filtered}, Q \right)
\end{equation}

\begin{table}[!h]
\centering
\caption{3D Semnatic Segementation on Replica Dataset.}
\label{tab:seg result}
\begin{tabular}{@{}cccc@{}}
\toprule
\multirow{2}{*}{\textbf{}}           & \multirow{2}{*}{\textbf{Method}}  & \multirow{2}{*}{\textbf{mAcc}} & \multirow{2}{*}{\textbf{F-mIoU}}   \\
                                     &                                   &                                &                                    \\ \midrule
\multirow{2}{*}{\textbf{Privileged}} & LSeg\cite{li2022language}                      & 33.39                          & 51.54                              \\
                                     & OpenSeg\cite{ghiasi2022scaling}                   & 41.19                          & 53.74                              \\ \midrule
\multirow{4}{*}{\textbf{Zero-shot}}  & MaskCLIP\cite{dong2023maskclip}                  & 4.53                           & 0.94                               \\
                                     & ConceptFusion\cite{conceptfusion}+ SAM\cite{ravi2024sam2} & 31.53                          & 38.70                              \\
                                     & ConceptGraphs\cite{gu2024conceptgraphs}             & \textbf{40.63}                 & 35.95                              \\
                                     & \textbf{Ours}               & 38.76      & \textbf{67.93} \\ \bottomrule
\end{tabular}
\end{table}

\section{Experiments}
\subsection{Datasets}
We evaluate our method on several widely used 3D datasets, including ScanRefer\cite{chen2020scanrefer}, Nr3D\cite{achlioptas2020referit_3d}, Sr3D\cite{achlioptas2020referit_3d}, AI2-THOR\cite{kolve2017ai2thor}, and Replica \cite{straub2019replica} datasets. 

\textbf{ScanRefer} In the ScanRefer\cite{chen2020scanrefer} dataset, we selected eight scenes from including living room, dining room, study, bedroom, conference room, bathroom, and other common household environments.

\textbf{Nr3d} For the Nr3D and Sr3D datasets\cite{achlioptas2020referit_3d}, we report metrics such as Overall, Easy, Hard, View-dependent, and View-independent. In the Nr3D dataset, we used queries constructed with natural language, reflecting the dataset's realistic scene characteristics. 

\textbf{AI2-THOR} The AI2-THOR dataset\cite{kolve2017ai2thor} is a widely used dataset for 3D scene understanding, containing diverse indoor environments. We utilized the provided 3D models and their corresponding annotations for our experiments.

\textbf{Replica} The Replica dataset\cite{straub2019replica} is a large-scale dataset for 3D scene understanding, containing diverse indoor environments. We utilized the provided 3D models and their corresponding annotations for our experiments.

\subsection{Implementation Details}
We applied the open-vocabulary detection model YoloWorld\cite{cheng2024yolo} for object detection, and a VLM-based image descriptor to extract objects appearing in the images. We use SAM2\cite{ravi2024sam2} for segmentation of the detection results. Additionally, we use SigLip\cite{zhai2023sigmoid} and DINOv2 \cite{oquab2023dinov2} as the text encoder and visual encoder, respectively. In the construction of DSM, all VLMs used are based on OpenAI’s GPT-4o-mini. During the fusion process, we set the visual threshold \( t_v \) as 0.4 , text threshold \( t_x \) as 0.8, geometric threshold \( t_v \) as 0.3,  and total threshold \( T \) as 1.5. In DSM-Grounding, we select \( k = 3 \) for the top-k relationships. 

\subsection{3D Semantic Segmentation of DSM}
\textbf{Evaluation Protocol}. To validate the foundational quality of our map, we first evaluate its open-vocabulary 3D semantic segmentation performance on the Replica dataset \cite{straub2019replica}. A key challenge in this zero-shot evaluation is aligning our VLM-generated, open-ended object descriptions with the dataset's fixed ground-truth class labels. To address this, we devise a systematic LLM-based mapping protocol. For each object in our DSM, we synthesize its rich semantic profile into a descriptive sentence using the template: \textit{This is \(o_{tag}\), its appearance attributes include \( a_a \), its physical attributes are \( a_p \), and its affordance attributes are \( a_o \).}. We then prompt an LLM to perform a classification task, selecting the most appropriate ground-truth label from the Replica dataset that corresponds to our generated description. This process programmatically assigns a class label to each object's point cloud, enabling a direct and fair comparison against the ground truth.

\textbf{Results and Analysis}. As presented in Table \ref{tab:seg result}, our method is benchmarked against both privileged, i.e., fine-tuned, and zero-shot baselines. We report mean Accuracy (mAcc) for overall scene segmentation and Foreground-mean IoU (F-mIoU) to specifically assess performance on foreground objects. Our DSM achieves an F-mIoU of 67.93, markedly surpassing all other methods, including the privileged OpenSeg approach, which scored 53.74. While our mAcc of 38.76 is competitive with the leading zero-shot method, the substantial lead in F-mIoU is particularly noteworthy. This superior performance on foreground objects is attributed to the rich, multi-dimensional attributes captured by our DSM. These diverse semantics furnish more discriminative features than simple class labels, thereby enabling the model to better distinguish between object instances and validating the efficacy of our foundational scene representation. 

\subsection{DSM-Grounding Experiment}
\textbf{Evaluation Protocol}. We evaluate DSM-Grounding on three standard 3D visual grounding benchmarks: ScanRefer \cite{chen2020scanrefer}, Sr3D \cite{achlioptas2020referit_3d}, and Nr3D \cite{achlioptas2020referit_3d}. For ScanRefer, following the standard protocol, we report Accuracy at IoU thresholds of 0.25 and 0.5 (Acc@0.25, Acc@0.5). For Sr3D and Nr3D, we report Top-1 accuracy. All evaluations are conducted under zero-shot settings and compared against state-of-the-art methods.

\begin{table*}[h]
\centering
\caption{Comparisons of 3D visual grounding on ScanRefer\cite{chen2020scanrefer} dataset. The Accuracy at 0.25 and 0.5 IoU thresholds is presented separately for “Unique,” “Multiple,” and “Overall” categories.}
\label{tab:scanrefer main}
\begin{tabular}{@{}cccccccccc@{}}
\toprule
\multirow{2}{*}{\textbf{Method}} &
  \multirow{2}{*}{\textbf{Venue}} &
  \multirow{2}{*}{\textbf{Supervision}} &
  \multirow{2}{*}{\textbf{LLMs}} &
  \multicolumn{2}{c}{\textbf{Unique}} &
  \multicolumn{2}{c}{\textbf{Multiple}} &
  \multicolumn{2}{c}{\textbf{Overall}} \\ \cmidrule(l){5-10} 
    &
   &
   &
   &
  \multicolumn{1}{l}{\textbf{Acc@0.25}} &
  \multicolumn{1}{l}{\textbf{Acc@0.5}} &
  \multicolumn{1}{l}{\textbf{Acc@0.25}} &
  \multicolumn{1}{l}{\textbf{Acc@0.5}} &
  \multicolumn{1}{l}{\textbf{Acc@0.25}} &
  \multicolumn{1}{l}{\textbf{Acc@0.5}} \\ \midrule
ScanRefer\cite{chen2020scanrefer}     & ECCV’20    & Fully       & -             & 67.60 & 46.20 & 32.10 & 21.30 & 39.00 & 26.10 \\
Scene-Verse\cite{jia2024sceneverse}   & ECCV’24    & Fully       & -             & 81.60 & 75.10 & 43.70 & 39.10 & 50.60 & 45.80 \\
LIBA\cite{wang2025liba}          & AAAI’25    & Fully       &   -            & 88.81 & 74.27 & 54.42 & 44.41 & 59.57 & 48.96 \\
\midrule
OpenScene \cite{peng2023openscene}    & CVPR’23    & Fine-tuning & CLIP          & 20.10 & 13.10 & 11.10 & 4.4   & 13.2  & 6.5   \\
Chat-3D v2    & NeurIPS’24 & Fine-tuning & Vicuna1.5-7B  & 61.20 & 57.60 & 25.20 & 22.6  & 35.9  & 30.4  \\
Inst3D-LMM\cite{yu2025inst3d-llm}    & CVPR’25    & Fine-tuning & Vicuna1.5-7B  & 88.60 & 81.50 & 48.70 & 43.20 & 57.80 & 51.60 \\ \midrule
ConceptGraphs\cite{gu2024conceptgraphs} & ICRA’24    & Zero-Shot   & GPT-4         & 16.50 & 10.32 & 9.57  & 7.69  & 13.28 & 9.31  \\
ZSVG3D\cite{yuan2024zsvg3d}        & CVPR’24    & Zero-Shot   & GPT-4 turbo   & 63.80 & 58.40 & 27.70 & 24.6  & 36.4  & 32.7  \\
VLM-Grounder\cite{xu2024Vlm-grounder}  & CoRL’24    & Zero-Shot   & GPT-4o        & 66.00 & 29.80 & 48.30 & 33.5  & 51.6  & 32.8  \\
SeeGround\cite{li2024seeground}     & CVPR’25    & Zero-Shot   & Qwen2-VL-72B  & 75.70 & 68.90 & 34.00 & 30.00 & 44.10 & 39.40 \\
FreeQ-Graph \cite{zhan2025freeq-graph}   & Arxiv'25          & Zero-Shot   & Qwen2-VL-72B  & 83.10 & 79.40 & 50.16 & 39.13 & 56.13 & 49.41 \\
\textbf{Ours} & -          & Zero-Shot   & GPT-4o-mini   & 83.32 & 80.17 & 47.01 & 43.93 & 57.47 & 55.39 \\
\textbf{Ours} & -          & Zero-Shot   & Qwen2.5-VL-7B & \textbf{88.57} & \textbf{88.57} & 51.46 & 48.54 & 59.38 & 57.19 \\
\textbf{Ours} &
  - &
  Zero-Shot &
  Qwen2.5-VL-72B &
  {85.71} &
  {85.71} &
  \textbf{56.98} &
  \textbf{53.65} &
  \textbf{61.56} &
  \textbf{59.06} \\ \bottomrule
\end{tabular}
\end{table*}

\begin{table*}[h]
\centering
\caption{Comparisons of 3D visual grounding on Sr3D [36] and Nr3D [36]. We evaluate the top-1 accuracy using ground-truth boxes. “super”: supervision method.}
\label{tab:Sr3d Nr3d}
\begin{tabular}{cccccccccccc}
\toprule
\multirow{2}{*}{\textbf{Method}} &
  \multirow{2}{*}{\textbf{Super}} &
  \multicolumn{5}{c}{\textbf{Nr3d}} &
  \multicolumn{5}{c}{\textbf{Sr3d}} \\ \cmidrule(l){3-12} 
 &
   &
  \textbf{Overall} &
  \textbf{Easy} &
  \textbf{Hard} &
  \textbf{V-Dep.} &
  \textbf{V-Indep.} &
  \textbf{Overall} &
  \textbf{Easy} &
  \textbf{Hard} &
  \textbf{V-Dep.} &
  \textbf{V-Indep} \\ \midrule
InstanceRefer \cite{yuan2021instancerefer} &
  Fully &
  38.80 &
  46.00 &
  31.80 &
  34.50 &
  41.90 &
  48.00 &
  51.10 &
  40.50 &
  45.80 &
  48.10 \\
LAR \cite{bakr2022lar} &
  Fully &
  48.90 &
  58.40 &
  42.30 &
  47.40 &
  52.10 &
  59.40 &
  63.00 &
  51.20 &
  50.00 &
  59.10 \\
MVT \cite{huang2022mvt} &
  Fully &
  59.50 &
  67.40 &
  52.70 &
  59.10 &
  60.30 &
  64.50 &
  66.90 &
  58.80 &
  58.40 &
  58.40 \\
ViL3DRel \cite{chen2022ViL3DRel}    &
  Fully &
  64.40 &
  70.20 &
  57.40 &
  62.00 &
  64.50 &
  72.80 &
  74.90 &
  67.90 &
  63.80 &
  73.20 \\
EDA \cite{wu2023eda} &
  Fully &
  52.10 &
  58.20 &
  46.10 &
  50.20 &
  53.10 &
  68.10 &
  70.30 &
  62.90 &
  54.10 &
  68.70 \\
3D-VisTA \cite{yang20243d-vista} &
  Fully &
  64.20 &
  72.10 &
  56.70 &
  61.50 &
  65.10 &
  76.40 &
  78.80 &
  71.30 &
  58.90 &
  77.30 \\
Scene-Verse \cite{jia2024sceneverse} &
  Fully &
  64.90 &
  72.50 &
  57.80 &
  56.90 &
  67.90 &
  77.50 &
  80.10 &
  71.60 &
  62.80 &
  78.20 \\
  \midrule
ZSVG3D \cite{yuan2024zsvg3d} &
  Zero-Shot &
  46.50 &
  31.70 &
  36.80 &
  40.00 &
  39.00 &
  - &
  - &
  - &
  - &
  - \\
VLM-Grounder \cite{xu2024Vlm-grounder} &
  Zero-Shot &
  48.00 &
  55.20 &
  39.50 &
  45.80 &
  49.40 &
  - &
  - &
  - &
  - &
  - \\
ConceptGraph \cite{gu2024conceptgraphs} &
  Zero-Shot &
  38.20 &
  39.40 &
  32.60 &
  42.10 &
  38.70 &
  43.60 &
  44.30 &
  41.90 &
  38.40 &
  49.70 \\

  SeeGround \cite{li2024seeground} &
  Zero-Shot &
  54.50 &
  38.30 &
  42.30 &
  48.20 &
  46.10 &
  65.40 &
  47.90 &
  52.20 &
  58.40 &
  56.20 \\
FreeQ-Graph \cite{zhan2025freeq-graph} &
  Zero-Shot &
  61.80 &
  61.40 &
  \textbf{57.80} &
  60.90 &
  \textbf{67.10} &
  70.90 &
  \textbf{79.30} &
  \textbf{63.90} &
  64.10 &
  \textbf{76.50} \\
Ours &
  Zero-Shot &
  \textbf{62.19} &
  \textbf{64.06} &
  56.00 &
  \textbf{61.25} &
  63.12 &
  \textbf{73.33} &
  77.44 &
  63.29 &
  \textbf{73.91} &
  73.49 \\ \bottomrule
\end{tabular}
\end{table*}

\textbf{Experiment Result.}
We evaluate our method against fully supervised, fine-tuned, and zero-shot approaches, with a primary focus on the zero-shot setting. As presented in Table \ref{tab:scanrefer main}, our method establishes state-of-the-art performance on the ScanRefer dataset. Specifically, when employing the Qwen2.5-VL-72B model, our approach secures the highest scores in both the \textit{Overall} category, achieving 61.56 Acc@0.25 and 59.06 Acc@0.5, and the challenging \textit{Multiple} category, with scores of 56.98 Acc@0.25 and 53.65 Acc@0.5. This performance surpasses all existing zero-shot methods, including the strong baseline from FreeQ-Graph \cite{zhan2025freeq-graph}. The results on the Sr3D and Nr3D datasets, detailed in Table \ref{tab:Sr3d Nr3d}, further corroborate the efficacy of our approach. On Nr3D, our method attains the best overall accuracy of 62.19, while on Sr3D, it achieves a leading overall accuracy of 73.33. These findings underscore the significant advantages of leveraging the rich contextual information from DSM across varying levels of descriptive complexity.

Furthermore, we validated our method's versatility by testing it with a range of VLMs, including GPT-4o-mini, Qwen2.5-VL-7B, and Qwen2.5-VL-72B. The consistently strong performance across these models, as detailed in Table \ref{tab:scanrefer main}, highlights the robustness and generalizability of the DSM-Grounding method.

\textbf{Ablation Study}
To dissect the contribution of each component, we performed ablation studies on the AI2-THOR dataset, with the results presented in Table \ref{tab:abli}. Our analysis focuses on the impact of the Latent Relation Filtering (LRF) module and the three semantic attributes: appearance ($a_a$), physical ($a_p$), and affordance ($a_o$). The results unequivocally demonstrate the efficacy of our complete model, which achieves the best overall performance with an Acc@0.25 of 61.59 and an Acc@0.5 of 60.00. The removal of the LRF module causes a substantial performance degradation of approximately 7-8 percentage points, most notably in the \textit{Multiple} category. This underscores the critical role of relational reasoning in object disambiguation. Further ablation of the semantic attributes indicates that while all three contribute positively, the appearance attribute ($a_a$) offers the most significant signal. This is evidenced by the relatively strong performance of the model even when physical and affordance attributes are excluded. Nevertheless, the combination of all diverse attributes and the LRF module is indispensable for attaining state-of-the-art results.

\begin{table*}[h] 
\centering
\caption{Grounding Result on ai2thor\cite{kolve2017ai2thor}}
\label{tab:grounding result ai2thor}
\begin{tabular}{@{}ccccccc@{}}
\toprule
\multirow{2}{*}{\textbf{Method}}    & \multicolumn{2}{c}{\textbf{Unique}}  & \multicolumn{2}{c}{\textbf{Multiple}} & \multicolumn{2}{c}{\textbf{Overall}} \\ \cmidrule(l){2-7} 
                                   & \textbf{Acc@0.25} & \textbf{Acc@0.5} & \textbf{Acc@0.25}  & \textbf{Acc@0.5} & \textbf{Acc@0.25} & \textbf{Acc@0.5} \\ \midrule
\multicolumn{1}{c|}{ZSVG3D\cite{yuan2024zsvg3d}}        & 11.11             & 9.72             & 19.08              & 12.92            & 17.63             & 12.34            \\
\multicolumn{1}{c|}{SeeGround\cite{li2024seeground}}     & \textbf{98.72}    & \textbf{98.65}   & 32.12              & 29.09            & 49.09             & 46.82            \\
\multicolumn{1}{c|}{\textbf{Ours}} & 98.67             & 98.32            & \textbf{48.79}     & \textbf{46.67}   & \textbf{61.59}    & \textbf{60.00}      \\ \bottomrule
\end{tabular}

\end{table*}

\begin{table*}[h]
\centering
\caption{Ablation Result on ai2thor\cite{kolve2017ai2thor}}
\label{tab:abli}
\begin{tabular}{@{}ccccccccccc@{}}
\toprule
\multirow{2}{*}{\textbf{}} &
  \multirow{2}{*}{\textbf{LRF}} &
  \multirow{2}{*}{\textbf{\begin{tabular}[c]{@{}c@{}}Appearance\\ Attribute\end{tabular}}} &
  \multirow{2}{*}{\textbf{\begin{tabular}[c]{@{}c@{}}Physics\\ Attribute\end{tabular}}} &
  \multirow{2}{*}{\textbf{\begin{tabular}[c]{@{}c@{}}Affordance\\ Attribute\end{tabular}}} &
  \multicolumn{2}{c}{\textbf{Unique}} &
  \multicolumn{2}{c}{\textbf{Multiple}} &
  \multicolumn{2}{c}{\textbf{Overall}} \\ \cmidrule(l){6-11} 
 &   &   &   &                        & \textbf{Acc@0.25} & \textbf{Acc@0.5} & \textbf{Acc@0.25} & \textbf{Acc@0.5} & \textbf{Acc@0.25} & \textbf{Acc@0.5} \\ \midrule
 & \Checkmark & \Checkmark & \Checkmark & \multicolumn{1}{c|}{\Checkmark} & 98.67             & 98.32            & \textbf{48.79}    & \textbf{46.67}   & \textbf{61.59}    & \textbf{60.00}      \\
 & \XSolidBrush & \Checkmark & \Checkmark & \multicolumn{1}{c|}{\Checkmark} & \textbf{99.53}    & \textbf{99.53}   & 39.7              & 38.18            & 54.77             & 53.64            \\
 & \XSolidBrush & \Checkmark & \XSolidBrush & \multicolumn{1}{c|}{\XSolidBrush} & 99.09             & 99.09            & 34.24             & 33.03            & 50.45             & 49.55            \\
 & \XSolidBrush & \XSolidBrush & \Checkmark & \multicolumn{1}{c|}{\XSolidBrush} & 98.18             & 98.18            & 35.76             & 32.73            & 51.36             & 49.09            \\
 & \XSolidBrush & \XSolidBrush & \XSolidBrush & \multicolumn{1}{c|}{\Checkmark} & 99.09             & 99.09            & 33.94             & 31.52            & 50.23             & 48.41            \\ \bottomrule
\end{tabular}
\end{table*}

\subsection{Robot Experiment}

To validate the practical applicability of our framework, we conducted experiments in both simulated and real-world robotic scenarios.

\textbf{Simulated Environment.} We evaluated our method in the high-fidelity AI2-THOR simulator \cite{kolve2017ai2thor}. As shown in Table \ref{tab:grounding result ai2thor}, our approach significantly outperforms existing zero-shot methods like ZSVG3D \cite{yuan2024zsvg3d} and SeeGround \cite{li2024seeground}. Notably, in the challenging \textit{Multiple} category, our method achieves an Acc@0.5 of 46.67, a substantial improvement over SeeGround's 29.09, highlighting the DSM framework's effectiveness in resolving ambiguity. These results, detailed further in Tabel \ref{tab:grounding result ai2thor}, confirm the robustness of our grounding pipeline in a controlled setting.

\begin{figure}[h]
    \centering
    \includegraphics[trim=220  100 180 20, clip, width=0.5\textwidth]{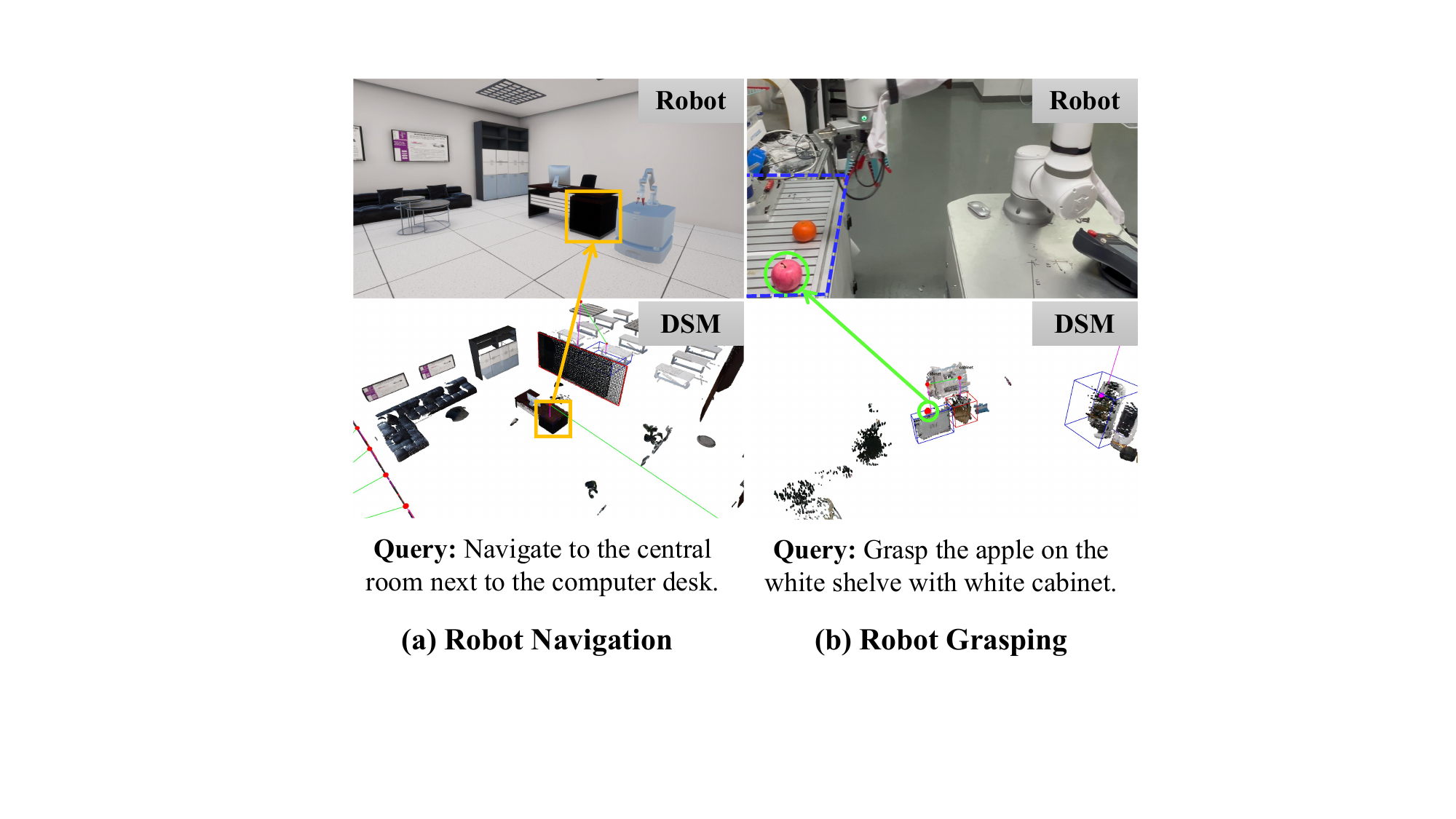}
    \caption{\textbf{Robot Experiment.}(a)Robot Navigation for blue book, (b)Robot Grasping for red apple}
    \label{fig:rob exp}
\end{figure}

\textbf{Real-World Deployment.}We deployed our system, DVS Platform\cite{zheng2025demonstratingdvsdynamicvirtualreal}, on a physical robot to perform a series of semantic navigation and grasping tasks in our laboratory. The robot first autonomously explored the environment to build a DSM. For navigation, given a command like \textit{Navigate to the central room next to the computer desk}, our system parsed \textit{central room} as the target destination and \textit{computer desk} as a key landmark. By DSN-Grounding, the robot localized the landmark and successfully navigated to the specified room. For grasping, a more complex command such as \textit{Grasp the apple on the white shelve with white cabinet} was issued. The DSM-Grounding module identified the target (\textit{apple}) and multiple anchors (\textit{white shelve}, \textit{white cabinet}). The Latent Relation Filtering (LRF) module then reasoned over the rich attribute and relational data in the DSM to precisely disambiguate the correct apple. Following successful localization, the robot navigated to the target and executed the grasp. These end-to-end demonstrations, illustrated in Figure \ref{fig:rob exp}, validate the practical utility and robustness of our framework in complex, real-world human-robot interaction scenarios.

\section{Conclusions}
In this work, we introduced the Diverse Semantic Map (DSM) framework, a novel scene representation framework that captures multi-dimensional attributes and relations, and DSM-Grounding, a method that transforms 3D visual grounding into a structured reasoning process. By leveraging the rich context of the DSM, our method establishes a new state-of-the-art in zero-shot 3D visual grounding on benchmarks like ScanRefer and significantly improves foreground semantic segmentation. While effective, the framework's performance is tied to the quality of upstream perception modules and the latency of large VLMs. Future work will focus on enhancing robustness through alternative 3D representations and improving real-time performance by exploring more efficient models, aiming to advance robotic adaptability in complex environments.


\bibliographystyle{IEEEtran}
\bibliography{root}

\end{document}